\documentclass[times, 10pt,twocolumn]{article} 
\usepackage{latex8}
\usepackage{times}
\usepackage{amsmath}
\usepackage{amsfonts}
\usepackage{array}
\usepackage{verbatim}
\usepackage{graphicx} 
\usepackage{float} 
\usepackage{subfigure} 
\usepackage{caption}

\begin{document}

\title{MER-GCN: Micro-Expression Recognition Based on Relation Modeling with Graph Convolutional Networks}

\author{Ling Lo\\
National Chiao Tung Univresity\\  lynn97.ee08g@nctu.edu.tw\\
\and
Hong-Xia Xie\\
National Chiao Tung Univresity\\ 
hongxiaxie.ee08g@nctu.edu.tw\\
\and
Hong-Han Shuai\\
National Chiao Tung Univresity\\ 
hhshuai@nctu.edu.tw\\
\and
Wen-Huang Cheng\\
National Chiao Tung Univresity\\ 
whcheng@nctu.edu.tw\\
}

\maketitle
\thispagestyle{empty}

\begin{abstract}
 Micro-Expression (ME) is the spontaneous, involuntary movement of a face that can reveal the true feeling. Recently, increasing researches have paid attention to this field combing deep learning techniques. Action units (AUs) are the fundamental actions reflecting the facial muscle movements and AU detection has been adopted by many researches to classify facial expressions. However, the time-consuming annotation process makes it difficult to correlate the combinations of AUs to specific emotion classes. Inspired by the nodes relationship building Graph Convolutional Networks (GCN), we propose an end-to-end AU-oriented graph classification network, namely MER-GCN, which uses 3D ConvNets to extract AU features and applies GCN layers to discover the dependency laying between AU nodes for ME categorization. To our best knowledge, this work is the first end-to-end architecture for Micro-Expression Recognition (MER) using AUs based GCN. The experimental results show that our approach outperforms CNN-based MER networks.
\end{abstract}

\vspace{-2em}
\Section{Introduction}
\vspace{-0.5em}
Micro Expression (ME) is a rapid, subtle and spontaneous motion of a human face that usually lasts between 1/25 and 1/5 seconds. Unlike macro-expression, which could be misleading on human emotion recognition, micro-expression is mostly expressed unconsciously where genuine emotion can be revealed. As indicators of emotional states, effective Micro-Expression Recognition (MER) can boost many practical applications in our daily life. For example, Ekman and Rosenberg defined six basic expressions \cite{ekman1997face}, namely happiness, sadness, anger, surprise, disgust and fear based on facial Action Units (AUs) with the facial action coding system (FACS).

\begin{figure}[ht]
\begin{center}
\includegraphics[scale = 0.9]{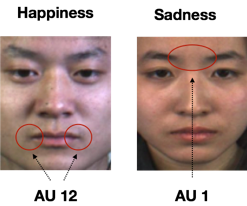}
\end{center}
\setlength{\abovecaptionskip}{-0.3cm}
\setlength{\belowcaptionskip}{-0.4cm}
\caption{AU regions illustration of “happiness” and “sadness” from the CASME II dataset \cite{casme}.}
\label{fig:AU}
\end{figure}

\begin{figure*} 
\centering 
\includegraphics[width=0.7\textwidth]{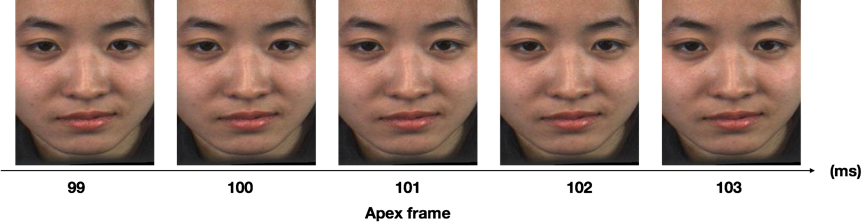} 
\setlength{\abovecaptionskip}{0cm}
\setlength{\belowcaptionskip}{-0.6cm}
\caption{Sample frames representing “happiness” from Subject 17 in CASME II dataset \cite{casme}. The apex frame shows the highest intensity of current expression.} 
\label{Casme_samples}
\end{figure*}

AU is an observable component of facial movement where distinct facial areas are correlating to fine-grained expression changes on faces. Distinct ME can be represented using different AUs combination. An illustration of AUs can be seen in \textbf{Fig~\ref{fig:AU}}. According to statistical calculation and facial anatomy information, strong relationships exist among different AUs under different facial expressions. Since the labeled classes in current ME datasets contain some biases, mapping extracted features to the corresponding emotion category is a tricky problem in the MER task (\textbf{Fig~\ref{Casme_samples}} shows some sample frames from MER dataset \cite{casme}). Although AUs cannot reflect the emotion classes directly, they can be regarded as a middle step towards high-efficiency MER task.

Since the graph is essentially a tool that can handle relationships between nodes, with the recent development of graph neural networks (GNN), e.g., Graph Convolutional Networks (GCN), relation modeling for visual tasks has attracted more and more attention \cite{mao2018hierarchical}\cite{liu2019relation}. GCN was first proposed for semi-supervised image classification \cite{GCN}. GCN based approaches are later developed for facial AU detection \cite{liu2019relation} but none has been found for MER.

Based on the above observations, we introduce an end-to-end AU-oriented architecture for MER, namely MER-GCN. Inspired by ML-GCN~\cite{ML-GCN}, after extracting feature representations by 3D ConvNets, we further apply GCN for AUs relation modeling. The proposed MER-GCN architecture can thus attain not only deep spatial-temporal features but also the information of hidden dependencies among the physical facial muscle movements, enabling accurate ME classification. The main contributions of this paper are twofold: (1) We provide a comprehensive survey to review existing solutions for MER, from CNN-based to GNN-based. (2) We propose an end-to-end AU-oriented network using GCN for MER. Experimental results demonstrate that our proposed framework performs the best on the benchmarking ME datasets.

The paper is organized as follows. Sec.~\ref{sec:related_works} gives related works. Sec.~\ref{sec:CNNMER} introduce CNN-based MER architecture in detail. Sec~\ref{sec:MERGCN} presents our proposed MER-GCN network. Sec.~\ref{sec:experiment} is the experiments and Sec.~\ref{sec:conclusion} makes conclusions and our future work.
\section{Related Works}
\label{sec:related_works}
\subsection{Low-level Feature Representations}
In the last decades, the ME feature extraction is mainly based on hand-crafted ways, e.g. LBP-TOP \cite{zhao2007dynamic} and HOOF \cite{chaudhry2009histograms}. While LBP-TOP and its variants are popular ways to extract low-level ME features, they are unable to recognize the motion of facial movements and AUs are hard to obtained based on LBP-TOP features. HOOF-based methods focus more on facial temporal dynamic changes. However, head-pose variations could affect MER results. 
\subsection{High-level Feature Representations }
Current challenges of MER lie in the environmental variation, spontaneous and subtle facial movements, and small datasets. Hand-crafted features above have the limitations in terms of the robustness and performance in terms of accuracy. End-to-end solutions based on CNN feature extraction for MER have thus been increased in recent years \cite{liong2018less}\cite{ranjan2017hyperface}\cite{tran2015learning}\cite{wang2017background}. Bi-Weighted Oriented Optical Flow (Bi-WOOF) \cite{liong2018less} was proposed to encode essential expressiveness using only the apex frame and the onset frame. More CNN-based MER methods will be introduced in Sec.~\ref{sec:CNNMER}.
\subsection{AU Detection}
Some researches believe that recognizing well-defined muscles in the face (i.e., AUs) can reduce bias, which is more optimal than discrete emotion categories. A number of AU detection models thus extract facial appearance textures \cite{zhao2007dynamic} and landmarks \cite{wang2013capturing}, based on general features in image processing tasks. By introducing deep learning to AU detection, models \cite{zhao2015joint}\cite{abousaleh2016novel}\cite{liu2013aware} can learn rich facial representations to capture action movements and achieve promising detection performance. However, these methods often overfit on a specific facial expression dataset since current datasets have a very limited number of annotated AUs. Labelling AU regions manually is extremely time-consuming work and requires professional annotators.

\subsection{Graph Convolutional Networks}
Considering videos as a hierarchical data structure, and the relations between frames are more complex than the order of a sequence, Mao {\em et al.} \cite{mao2018hierarchical} modeled the video frames by a deep convolution graph network (DCGN). Liu {\em et al.} \cite{liu2019relation} used GCN for AU relation modeling, which has not been explored before. Latent representations of extracted AU-related regions are learned through an auto-encoder, and AU relationships are modeled through GCN.
\section{CNN-Based MER Architecture}
\label{sec:CNNMER}
\begin{figure*} 
\centering 
\includegraphics[width=0.7\textwidth]{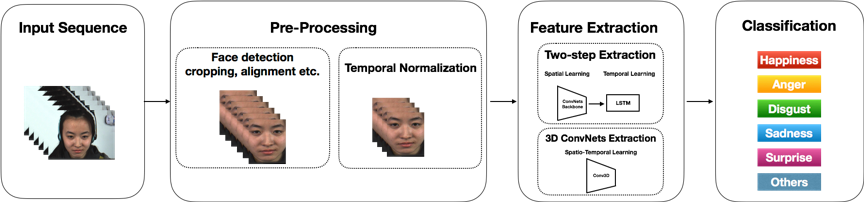} 
\setlength{\abovecaptionskip}{0cm}
\setlength{\belowcaptionskip}{-0.5cm}
\caption{The general pipeline of CNN-based MER task.} 
\label{CNN-MER} 
\end{figure*}
The CNN-based MER architectures are summarized in \textbf{Fig~\ref{CNN-MER}}. It usually contains three main steps: pre-processing, feature extraction and ME classification. The input video frames with subtle and rapid facial movements are often collected by a high-speed camera. These facial images are then cropped and pre-processed for expression recognition. In recent years, CNN-based landmark localization methods have been proposed \cite{deng2018facial}\cite{ranjan2017hyperface}\cite{pfister2011recognising}\cite{bargal2016emotion}. Since the number of frames in each sequence varies, temporal normalization approaches have been considered to generate fixed-length normalized frames. Pfister {\em et al.} \cite{pfister2011recognising} proposed a popular video normalization algorithm, temporal interpolation model (TIM), to convert frames from the constructed data manifold into a time-limited sequence. For feature-level frame aggregation, the learned features of frames in the sequence are aggregated.

\textbf{Feature Extraction: Two-step learning} Since MEs are only temporally exhibited in a fraction of a second and involve minute spatial changes. There are two main spatio-temporal feature extraction architectures for MER. The first one is a two-step, sequential model, where typical CNNs extract spatial feature representations among all frames and Long Short-Term Memory (LSTM) based Recurrent Neural Network (RNN) for temporal correlation exploring between the frames \cite{wang2018micro}. Enriched Long-term Recurrent Convolutional Network (ELRCN) \cite{khor2018enriched} embeds two learning modules: Spatial Dimension Enrichment and Temporal Dimension Enrichment. Also, the optical flow was used in this work to enrich the input data. However, this kind of model has been proved worse than 3D convolutional neural networks (3D ConvNets) to learn the spatio-temporal relationship \cite{reddy2019spontaneous}.

\textbf{Feature Extraction: 3D ConvNets} Happy and Routray \cite{happy2017fuzzy} argued that the changes on the face during a ME is temporal changes more than spatial. Using 3D ConvNets \cite{tran2015learning} is an alternative way for ME feature extraction. 3D ConvNets are an extension of 2D CNN. Liu {\em et al.}~\cite{liu2014deeply} manually defined 13 facial parts and used 3D filters to convolve feature maps for facial action part detection. There are two types of 3D-CNN models for spontaneous MER in \cite{reddy2019spontaneous}, i.e., MicroExpSTCNN and MicroExpFuseNet, by exploiting the spatio-temporal information in the CNN framework. The former model considers the full-face input, whereas the latter receives a fusion of the eye and mouth regions as input.
\section{GNN-Based MER Architecture}
\label{sec:MERGCN}
The labeled data for a micro-expression recognition task sometimes contain potential bias due to the recognition difficulties when labeling by the human annotators, and thus the mapping procedure becomes tricky even with a well-trained CNN to obtain effective frame features.
On the other hand, observable components like AUs aim to analyze the physical movement of the facial muscles, and hence are relatively objective. 
Taking the advantage of that, we introduce an AU-oriented architecture to recognize the micro-expressions based on the hypothesis that facial muscle movement is consecutive. Inspired by \cite{ML-GCN}, our work called MER-GCN applies a GCN on top of the Conv3D architecture mentioned above in order to explore the dependencies among different AUs. 
The proposed CNN-GCN architecture can attain not only deep spatial-temporal image features but also the hidden inter-relations among the physical muscle movements on faces. 

\subsection{Convolution on Graph}
Just like CNN can capture the most significant information within pixels in images, a graph-based learning algorithm targets on learning the relation between each object node from the non-Euclidean data described in the form of graph. As standard convolution computation can be seen as learning kernels to discover meaningful and distinguishable latent patterns in images, graph convolution computation learns about the latent embedded nodes based on the neighbors and the graph relations. Therefore, as standard convolutions are performed on local regions in Euclidean structure data, what is done in a graph convolution network can be seen as passing data through different nodes and the goal is to learn a function \(f\) that can update the node representations layer by layer.

A GCN takes the node feature description $X \in   \mathbb{R}^ {d \times n}$ and adjacency matrix $A \in   \mathbb{R}^ {n \times n}$  as input (where $n$ represents the nummber of nodes and $d$ stands for the dimension of feature description of each node), and uses the convolutional propagation function $f$ to update the $l$-{\em th} hidden layer $H^{l}$, where $H^{0} = X$. The outputs of $L$-layer GCN are embedded nodes represented by the last hidden layer $H^{L}$. Generally, each graph convolutional layer can be written as:
\vspace{-0.5em}
\begin{equation}
\vspace{-0.5em}
H^{l} = f(H^{l-1} \cdot A)\label{eq:1}
\end{equation}
Since $f$ is a propagation function \cite{GCN}, Eqn.~(\ref{eq:1}) can be further extended to:
\vspace{-0.5em}
\begin{equation}
\vspace{-0.5em}
H^{l} = \sigma(A H^{l-1} W^{l-1})\label{eq:2}
\end{equation}
where $\sigma$ is the non-linear activation function and $W^{l-1} \in \mathbb{R}^ {d \times d'}$ is the $(l-1){th}$ weighted matrix as $d$ and $d'$ stand for the input and output dimension of layer $l$, respectively.
The graph convolution operation can also be stacked to multiple layers as similar to the standard convolution. A stacked GCN model is able to learn the node dependencies after a few iterations.

\begin{figure}[ht]
\setlength{\belowcaptionskip}{-0.4cm}
\begin{center}
\includegraphics[width=\linewidth]{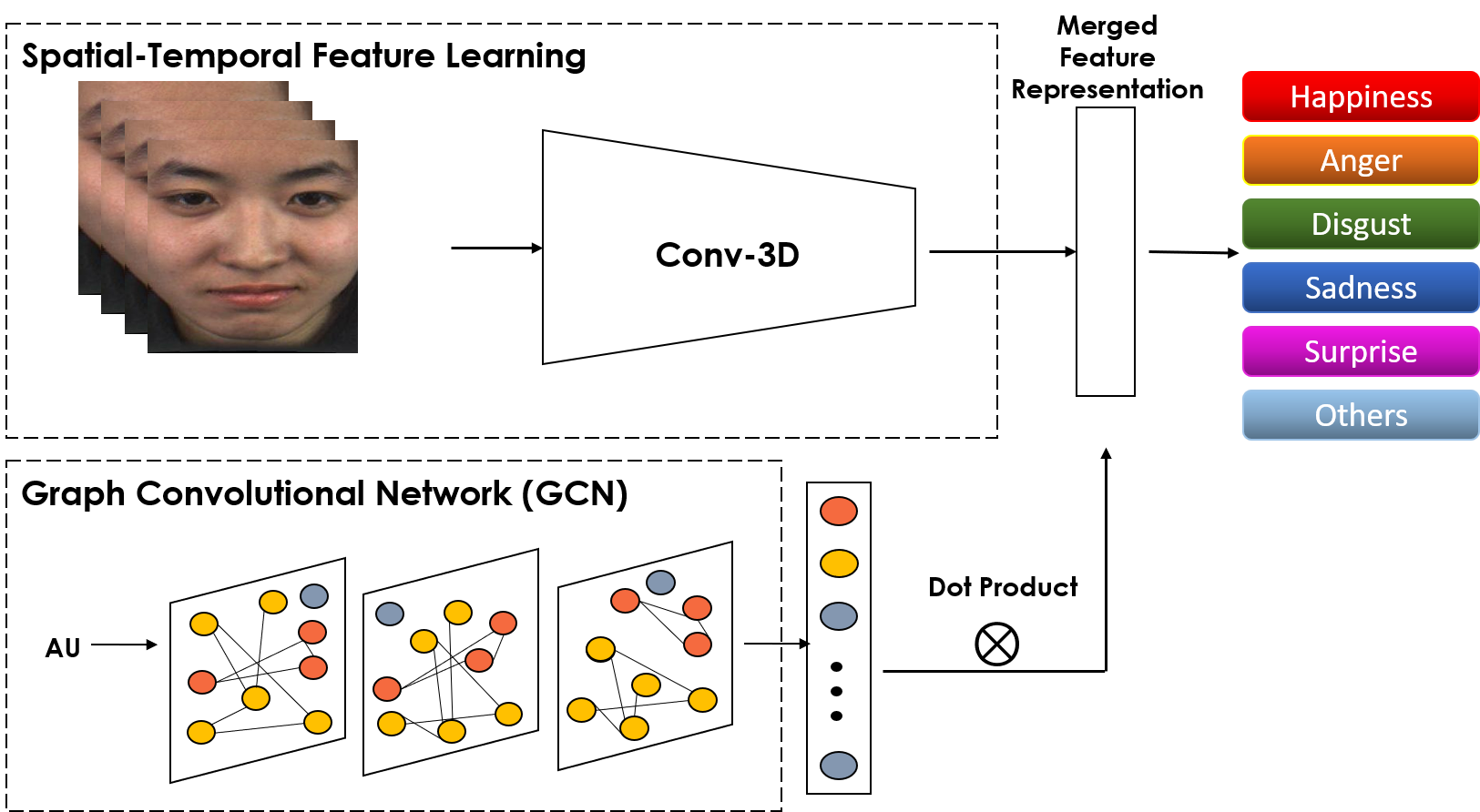}
\end{center}
\setlength{\abovecaptionskip}{-0.3cm}
\setlength{\belowcaptionskip}{-0.4cm}
\caption{Overview of our proposed MER-GCN framework.}
\label{fig:MER_GCN}
\end{figure}

\subsection{MER-GCN Approach}
\textbf{Building graph}  
As the first proposed GCN in \cite{GCN} designed the final output as the prediction score of each node for semi-supervised classification, we design our final output layer in GCN to learn embedding the informative representation of AUs by the propagation function $f$.

Building the adjacency matrix $A$ is the significant first step in constructing GCN. In this paper, owing to the fact that there is no well-defined correlation matrix describing the AU dependencies, we build the adjacency matrix by a data-driven way, using the co-occurrence of each pair of AUs in the train dataset as the inter-relation. 
Specifically, we model the co-occurrence by the conditional probability $P(U_i|U_j)$ to denote the chance of the $i$-th AU, $U_{i}$, co-occurring with the $j$-th AU, $U_{j}$. Assuming we have $n$ nodes in our graph, we first count the occurrence of each pair of the nodes in the training data set, and divide the concurring amount of $U_{i}$ and $U_{j}$ by the total occurrence of the $i$-th AU, $N_{j}$, to denote the conditional probability $P(U_i|U_j)$. The element $A_{ij}$ in adjacency matrix can be formulated by
\vspace{-0.5em}
\begin{equation}
\vspace{-0.5em}
A_{ij} = P(U_i|U_j) = \frac{N_{i\cap j}}{N_{j}}\label{eq:3}
\end{equation}
Note that $P(U_i|U_j)$ does not share the same value with $P(U_j|U_i)$, which results in the asymmetrical adjacency matrix. 
Another key factor is the feature representation of each node at the very beginning. As the name of AUs does not have a semantic meaningful structure, here we transform every AUs to the machine-comprehensible labels by one-hot encoder. The key idea is that the representation of every AUs is equally dependent to each other at first, and the output GCN is designed to learn the meaningful node representation as the ultimate output.

\textbf{3D CNN backbone}
In our proposed method, we first use a 3D ConvNet architecture to exploit the spatial-temporal features of frames. Our backbone model for sequence level feature extraction is based on 3D-Resnet-18 architecture \cite{3DCNN}. We obtain the output after several residual blocks as the spatial-temporal features. Then, we employ a global space-time pooling to acquire the feature map in a proper size. Since the convolution in this architecture is operated over both space and time dimensions collaboratively, the network can be trained to preserve and propagate both spatial and temporal information simultaneously, and thus catch more representative features than working separately in a two-step network.

\textbf{GCN based MER} 
To obtain the hidden knowledge among different AUs, we use a stacked-GCN to attain the unified representation $H^L$ from the original one-hot encoded node inputs $X$. After that, we apply the learned embedded vectors of AU representation to our sequence level feature extracted by CNN. The resulted vector is then fed into a fully-connected layer to form the final recognition result. The whole architecture is shown in \textbf{Fig~\ref{fig:MER_GCN}}.
While training, $W$ in Eqn.~(\ref{eq:2}) that stands for weighting matrix is the only trainable variable set in the GCN and is co-trained with the 3D-CNN architecture with the classification loss of cross entropy $L$ as follows:
\vspace{-1em}
\begin{equation}
\vspace{-0.5em}
L = -\sum_{c=1}^{n}y_{o,c}\log P_{o,c}
\label{eq:4}
\end{equation}
where $y_{o,c}$ is the binary indicator if label $c$ is same as the observation result ${o}$, and $P_{o,c}$ is the predicted probability observation result ${o}$ of label $c$.
\begin{table}[]
\setlength{\belowcaptionskip}{-0.7cm}
\caption{Our backbone R3D architecture used in the experiment. The size comes in order of $T,H,W$ and stands for the length, height and width of sequences. For every residual block, the number followed by the kernel size is the total number used in each layer. Besides there is the number of basic blocks in a residual block. }
\begin{tabular}{c|c|c}
layer                & kernel size              & output size \\ \hline
conv\_1              & \multicolumn{1}{c|}{$3\times7\times7, 64$, stride $1\times2\times2$}                              & $T\times56\times56$ \\ \hline

res\_block\_1  & 
\begin{minipage}{3.5cm}
\centering

$\begin{bmatrix}
3\times3\times3, & 64\\
3\times3\times3, & 64
\end{bmatrix} \times 2$\\

\end{minipage}  
& \multicolumn{1}{c}{$T\times56\times56$} \\
\hline

res\_block\_2 &
\begin{minipage}{3.5cm}
\centering

$\begin{bmatrix}
3\times3\times3, & 128\\
3\times3\times3, & 128
\end{bmatrix} \times 2$\\

\end{minipage}
& \multicolumn{1}{c}{
$\frac{T}{2}\times28\times28$} \\
\hline

res\_block\_3 & 
\begin{minipage}{3.5cm}
\vspace{1mm}
\centering

$\begin{bmatrix}
3\times3\times3, & 256\\
3\times3\times3, & 256
\end{bmatrix} \times 2$

\end{minipage}
& \multicolumn{1}{c}{
$\frac{T}{4}\times14\times14$} \\
\hline

res\_block\_4 & 
\begin{minipage}{3.5cm}
\vspace{1mm}
\centering

$\begin{bmatrix}
3\times3\times3, & 512\\
3\times3\times3, & 512\\
\end{bmatrix} \times 2$

\end{minipage}
& \multicolumn{1}{c}{
$\frac{T}{8}\times7\times7$} \\
\hline
\multicolumn{2}{c|}{global average pooling} & \multicolumn{1}{c}{
$512\times1$
} \\
\end{tabular}
\label{tab:backbone}
\end{table}
\section{Experiments}
\label{sec:experiment}
\subsection{Datasets and Experimental Settings}
\label{sec:dataset}
Public datasets for MER can be divided into non-spontaneous and spontaneous. Considering the fact that limited number of non-spontaneous datasets is insufficient for MER analysis, and the spontaneous nature of ME, in this work we use a spontaneous one, CASME II (Chinese Academy of Sciences Micro-Expression II)~\cite{casme} for experiment.
CASME II is an improved version of CASME collected by Yan {\em et al.} in well-controlled lab environment. It contains 247 micro-expressions (35 subjects) from over $3,000$ facial movements which were labeled based on AUs, participants’ self-report and the content of the stimuli video. However, CASME II suffers severe class unbalance since micro-expressions in this dataset are defined in 7 classes which are happiness (33 samples), disgust (50 samples), surprise (25 samples), repression (27 samples) and others (102 samples).

All the experiments were conducted on a workstation running Ubuntu 16.04 with 3.2GHz CPU, 64GB RAM, and NVIDIA GeForce GTX 2080 Ti GPU. We use Pytorch for the network implementation.

\subsection{Implementation Details}
\textbf{Training} In our experiment, we took a 18 layers 3D-Resnet architecture \cite{3DCNN} as our backbone model for frame level feature extraction. Owing to the fact that applied training data set is rather small, the networks is trained with undefined length of sequences by taking single sequence as input during every batch learning. As shown in Table~\ref{tab:backbone}, the model contains five residual blocks in total. The input is a sequence of length $T$, with each image resized to 112$\times$112 resolution and the pixel values are normalized. After the space-time dimensional convolution, we then use a global pooling layer to down-sampling our feature map into the size of $512\times1$ for further classification. 
Next, we set our GCN to have two stacked layers, as the dimension of each layer output is $1,024$ and $512$, respectively. We perform dot product operation to merge the sequence level features and the AU level features, and apply a fully-connected layer for the overall classification.

\textbf{Validation}
For validation, we use two different ways to cross validate our model: leave-one-subdect-out (LOSO) and k-folds cross validation.
For LOSO, we randomly leave all data from a single subject for validation during every run of training to lower the chance of having subject bias. For k-folds validation, we divide the dataset into k portions and randomly use one for validation at different runs of training. Both of the cross validation strategies can prevent our model from having different kinds of bias. Further comparison details will be discussed in Sec.~\ref{sec:cv}.

\subsection{Experimental Results}
\label{sec:cv}
Table~\ref{table:acc} shows the overall result of our experiment, where we evaluate the performances in the term of accuracy. The left subtable are the results using k-fold validation while the right subtable are the results of LOSO validation.
The 3D-CNN pre-trained model on the first row is using 3D-Resnet model as the backbone with pretrained weight for the 5 residual blocks on kinetics dataset~\cite{kinetics}. For both experiments, the model with pre-trained weight has the lowest accuracy due to the reason that kinetics dataset is a human action dataset, therefore the 3D-CNN part may not capture the useful features on a MER dataset. The second row of both tables are the results of a 18-layer 3D-Resnet. After training, the 3D-CNN architecture is able to model the spatial-temporal relation by convolution operation within both dimensions. As a result, a well-trained 3D-CNN model can achieve a better performance. 
Comparing to other deep learning based methods, our proposed approach appears to achieve the highest accuracy of 58.82 and 42.71 using k-fold and LOSO validation, respectively. Using the same backbone architecture with the previous 3D-CNN model, the progress on the accuracy shows that by joining the AU dependent information learned in GCN, our model further learn the relation between facial physical muscle movements and different micro-expressions and thus understand the micro-expression deeper and better. 

\textbf{Discussion} We observe from the two tables and notice that using k-fold cross validation strategy will obtain a better accuracy than using LOSO. One possible reason is due to the bias existing in CASME II dataset. As we mentioned in Sec.~\ref{sec:dataset}, the datasets we have in the field of MER suffers severe class unbalance. That is, for every subject, they do not have the same amount of data in different categories. Consequently, the experimental result using LOSO validation may suffer from the class bias while we are trying to avoid subject bias.

\textbf{Challenge} Based on our observations, the problem of MER is still very challenging when applying deep learning base architectures. When most of the deep learning models rely on a large amount of data to discover the general patterns in between, the limited size of existing MER datasets are very likely to cause over-fitting issue. Another challenge is the severely biased dataset. Our model has difficulty in recognizing the categories that have few samples in the training dataset as well as most of the existing MER approaches, since the model tends to lower the overall loss by avoiding rare types.

\begin{table}[]

\setlength{\belowcaptionskip}{-1cm}
\caption{Comparison of the accuracy using different validation methods.}
\begin{minipage}{0.48\linewidth}
\centering
\begin{tabular}{c|c}
Model & Acc(\%) \\ \hline
\begin{tabular}[c]{@{}c@{}}3D-CNN \\ (pre-trained)\end{tabular} & 52.37 \\ \hline
3D-CNN & 54.05 \\ \hline
MER-GCN & \textbf{58.82}
\end{tabular}\\
(a) k-fold
\end{minipage}
\begin{minipage}{0.48\linewidth}
\centering
\begin{tabular}{c|c}
Model & Acc(\%) \\ \hline
\begin{tabular}[c]{@{}c@{}}3D-CNN \\ (pre-trained)\end{tabular} & 39.64 \\ \hline
3D-CNN & 42.03 \\ \hline
MER-GCN & \textbf{42.71}
\end{tabular}\\
(b) LOSO
\end{minipage}
 \label{table:acc}
\end{table}

\section{Conclusions and Future Work}
\label{sec:conclusion}
In this paper, we investigate the field of MER especially using deep learning architectures, and propose our MER-GCN architecture which is, to our best knowledge, the first end-to-end MER system. Combining CNN and GCN allows our work to recognize sequence level feature as well as AU level feature in order to avoid the labeling bias.
Our experiments show that the additional information carried by stacked-GCN can contribute to MER and reach a higher accuracy.
We also notice that due to the class bias issue and the limit size of existing dataset, MER using deep learning based methods are confronted with the over-fitting issue. In the future, we can extend our work by augmenting data by synthetic strategy to enlarge the size of dataset for building a deeper and more complex model.

\section{Acknowledgement}
\label{sec:ack}
This work was supported in part by Ministry of Science and Technology of Taiwan under grants:  MOST 108-2218-E-009-056, MOST 108-2745-8-009-002, MOST  108-2634-F-007-009, MOST 108-2823-8-002-004, MOST 108-2218-E-002-055, and MOST 107-2221-E-182-025-MY2. 

\bibliographystyle{latex8}
\bibliography{latex8}

\end{document}